\pdfoutput=1

\documentclass[conference]{IEEEtran}
\IEEEoverridecommandlockouts
\usepackage{cite}
\usepackage{amsmath,amssymb,amsfonts}
\usepackage{algorithmic}
\usepackage[ruled,linesnumbered]{algorithm2e}
\usepackage{graphicx}
\usepackage{textcomp}
\usepackage{xcolor}
\def\BibTeX{{\rm B\kern-.05em{\sc i\kern-.025em b}\kern-.08em
    T\kern-.1667em\lower.7ex\hbox{E}\kern-.125emX}}
    
\renewcommand\baselinestretch{1}

\begin{document}

\title{Improving DNN Fault Tolerance using Weight Pruning and Differential Crossbar Mapping for ReRAM-based Edge AI}

\author{
    Geng Yuan\text{$^\star$}\textsuperscript{\rm 1}\thanks{$^\star$These Authors contributed equally.},
    Zhiheng Liao\text{$^\star$}\textsuperscript{\rm 2},
    Xiaolong Ma\textsuperscript{\rm 1},
    Yuxuan Cai\textsuperscript{\rm 1},
    Zhenglun Kong\textsuperscript{\rm 1},
    Xuan Shen\textsuperscript{\rm 1},
    Jingyan Fu\textsuperscript{\rm 2},\\
    Zhengang Li\textsuperscript{\rm 1},
    Chengming Zhang\textsuperscript{\rm 3},
    Hongwu Peng\textsuperscript{\rm 4},
    Ning Liu\textsuperscript{\rm 1},
    Ao Ren\textsuperscript{\rm 5},
    Jinhui Wang\textsuperscript{\rm 6},
    Yanzhi Wang\textsuperscript{\rm 1}\\
    \itshape\small
    \textsuperscript{1}Northeastern University, 
    \textsuperscript{2}North Dakota State University, 
    \textsuperscript{3}Washington State University,
    \textsuperscript{4}University of Connecticut,\\
    \itshape\small
    \textsuperscript{5}Chongqing University,
    \textsuperscript{6}University of South Alabama\\
    {\tt\small \textsuperscript{\rm 1}\{yuan.geng, ma.xiaol, cai.yuxu, kong.zhe, shen.xu, li.zhen, yanz.wang\}@northeastern.edu,} \\
    {\tt\small \textsuperscript{\rm 2}\{zhiheng.liao, jingyan.fu\}@ndsu.edu,}
     {\tt\small \textsuperscript{\rm 3}chengming.zhang@wsu.edu,}
    {\tt\small \textsuperscript{\rm 4}hongwu.peng@uconn.edu,} \\
    {\tt\small\textsuperscript{\rm 5}ren.ao@cqu.edu.cn, }
    {\tt\small \textsuperscript{\rm 6}jwang@southalabama.edu}
    
}

\maketitle

\begin{abstract}
Recent research demonstrated the promise of using resistive random access memory (ReRAM) as an emerging technology to perform inherently parallel analog domain in-situ matrix-vector multiplication---the intensive and key computation in deep neural networks (DNNs).
However, hardware failure, such as stuck-at-fault defects, is one of the main concerns that impedes the ReRAM devices to be a feasible solution for real implementations.
The existing solutions to address this issue usually require an optimization to be conducted for each individual device, which is impractical for mass-produced products (e.g., IoT devices).
In this paper, we rethink the value of weight pruning in ReRAM-based DNN design from the perspective of model fault tolerance.
And a differential mapping scheme is proposed to improve the fault tolerance under a high stuck-on fault rate.
Our method can tolerate almost an order of magnitude higher failure rate than the traditional two-column method in representative DNN tasks. 
More importantly, our method does not require extra hardware cost compared to the traditional two-column mapping scheme.
The improvement is universal and does not require the optimization process for each individual device.

\end{abstract}
\section{Introduction}
DNNs have achieved significant progress in many fields such as computer vision~\cite{goodfellow2016deep}, natural language processing~\cite{devlin2018bert}, and medical diagnosis~\cite{medical_iccad}. It becomes a fundamental building block of many artificial intelligence (AI) applications, such as facilitating edge devices in the Internet of Things (IoT) system.
However, to pursue better performance (i.e., accuracy), the ever-growing size and computation load make the DNN difficulty to be deployed on edge devices in IoT system, since edge devices usually have limited resources and require ultra-low power consumption. 

ReRAM enabled in-memory computing, as one of the emerging technologies, has shown advantages such as near-zero leakage-power, high active-power efficiency, non-volatility, and concise circuit structure. Accordingly, it is especially suitable for edge devices in IoT. ReRAM cells can be used to form a crossbar structure to conduct in-situ dot products, such as the convolution computations in DNNs.
However, one key challenge that prevents the ReRAM from becoming a practical technology and used in real hardware implementations is the reliability issue.
For example, the stuck-at-fault defect will lead the ReRAM cell's conductance to be fixed to a high or low conductance state, regardless of its programmed, as known as the stuck-on fault or stuck-off fault.
For a DNN implementation, when mapping the well-trained DNN model to the ReRAM crossbars, the stuck-at-fault defects result in a mismatch between target weights and the actual values mapped onto the crossbar, and eventually lead to an undesirable accuracy drop.

Several methods are proposed to address the stuck-at-fault defects, such as permuting the order of the crossbar rows and columns~\cite{chen2017accelerator}, retraining the network by considering the defect locations~\cite{liu2017rescuing, xia2017fault}.
These methods require individual optimization process for each ReRAM crossbar, and may also introduce complex control circuits, resulting in additional hardware overhead.
Although these methods are effective in mitigating the accuracy drop caused by the stuck-at-fault defects, even~\cite{deliang_handling} can restore 99\% of the accuracy drop, but for mass-produced IoT products, applying optimization for each individual product (IoT device) will bring a huge time cost, and is not realistic.
Thus, it is desirable to have a universal approach to improve the DNN fault tolerance.

DNN weight pruning technique is another active research area that has been intensively studied.
It can effectively reduce the number of DNN weights as well as the computations, achieving high-efficiency DNN inference.
Especially, the unstructured pruning allows the weights at arbitrary locations to be pruned~\cite{ren2018admmnn}, which achieves relative low accuracy degradation under a high pruning ratio.

Although the weight pruning techniques are used for compressing the DNN model to save hardware and speed up the inference, it is still considered as a trade-off between model accuracy and hardware costs. However, the essence of weight pruning is to zero out the redundant weights in the DNNs. This provides a unique possibility to improve the tolerance of DNNs to the stuck-off fault because the pruned weight is possible to fall exactly on the ReRAM crossbar stuck-off fault location, avoiding the mismatch between target weight and the actual value represented by the ReRAM cell.

In this paper, we rethink the value of weight pruning in ReRAM-based DNN design from the perspective of model fault tolerance.
We argue that weight pruning is no longer just a technique that sacrifices accuracy in exchange for hardware reduction. Unstructured weight pruning can also be used as a technique to improve model fault tolerance, meanwhile resulting in improved DNN accuracy on ReRAM-based design.

Our experimental results demonstrate that lower than a certain pruning ratio, the ability of model fault tolerance increases as the pruning ratio goes up. 
After exceeding a certain pruning ratio, the model fault tolerance decreases again.
The pruning can only improve the model fault tolerance to the stuck-off fault, but as reported in~\cite{failure_model}, the failure rate of stuck-on fault can be 5.2$\times$ higher than the stuck-off fault and become the major factor for the degradation of accuracy.
Thus, we propose a differential mapping scheme to mitigate the impact of stuck-on fault and improve the overall accuracy without additional hardware costs.

The contributions of our paper can be summarized as:
\begin{itemize}
    \item We propose to utilize unstructured pruning
    to improve fault tolerance of the ReRAM-based DNN in the edge AI. 
    The improvement is statistically achieved for all models, which avoids conducting optimizations for each individual device.
    \item We explore the relation between model sparsity and its robustness and propose a hierarchical progressive pruning method, which can find the best-suited pruning ratio for DNNs fault tolerance. 
    \item We propose to use a differential mapping scheme to convert the stuck-off fault tolerance achieved by pruning to stuck-on fault tolerance, which can effectively improve the overall fault tolerance when the stuck-on fault is the major fault defect. 
    \item We validate the effectiveness of proposed methods on potential tasks for IoT applications, including image classification and object detection.
\end{itemize}

\section{Background}

\subsection{ReRAM Crossbar and ReRAM-based DNN Accelerator}

Resistive RAM (ReRAM), as an emerging memory device, has many promising features, including non-volatility, nearly zero leakage power, high integration density, and high scalability~\cite{chua1971memristor,8053125}. 
Every memory cell comprises a ReRAM element to store one or more data bits. 
In addition to data storage, its memory array structure, as known as the ReRAM-based crossbar, is capable of performing in-situ dot product between an input vector (i.e., input feature maps) and stored numbers (i.e., weights).

For DNNs, the hardware accelerators have been intensively studied in recent years~\cite{ding2017circnn, HW1, wang2018towards, HW2, 8357306,10.1145/3394885.3431627, HW3, HW4, HW5}.
Specifically,
specialized hardware acceleration using ReRAM-based crossbars has been studied in recent years. 
Previous work, such as PRIME~\cite{PRIME} and ISAAC~\cite{shafiee2016isaac}, leverages in-situ computation to avoid the tremendous cost of data movement and efficiently compute multiply-accumulate operations, the most intensive computation in DNNs.
Recent work, TIMELY~\cite{li2020timely} saves the energy cost of data movements and D/A and A/D domain conversion by enhancing analog data locality to keep computations in analog domain.

\subsection{Hardware Failure}
Similar to the other nano-scale devices, hardware failure that greatly degrade the accuracy of neural networks perplexes the application of ReRAM \cite{chen2017accelerator, liu2017rescuing}.  
Some techniques have been proposed to model and detect the faults in ReRAM crossbar array including fault model with testing scheme \cite{kannan2015modeling}, a march algorithm to cover defect \cite{chen2014rram}, and analyzing impacts of stuck-at-fault defects on accuracy of a sparse coding network \cite{sheridan2015defect}. 
Various schemes to tolerate faults in ReRAM crossbar array have been proposed in some prior works. 
In hardware level, isolating faulty ReRAM is proposed by switching off access transistor \cite{manem2010design}. However, for the large ReRAM crossbar array implemented in complicated neural networks, tremendous routing and area overheads are inevitable, utilizing controlling individual transistor access. 
In \cite{liu2017rescuing}, redundant columns of ReRAM crossbars are utilized as a substitution for the defect ReRAM columns. But this method not only introduces nontrivial area overhead, but also increases design complexity of peripheral circuit. 
Unlike hardware level schemes, fault-aware network retraining with weight mapping/remapping are commonly composed in software level optimizations \cite{chen2017accelerator, liu2017rescuing, xia2017fault}.
They retrain the neural network by forcing the weights to be the same as the fault value on corresponding crossbar locations.
However, the retraining and remapping will introduce tremendous computing and interaction costs.

\subsection{DNN Weight Pruning}
\begin{figure} [t]
     \centering
     \includegraphics[width=0.9\columnwidth]{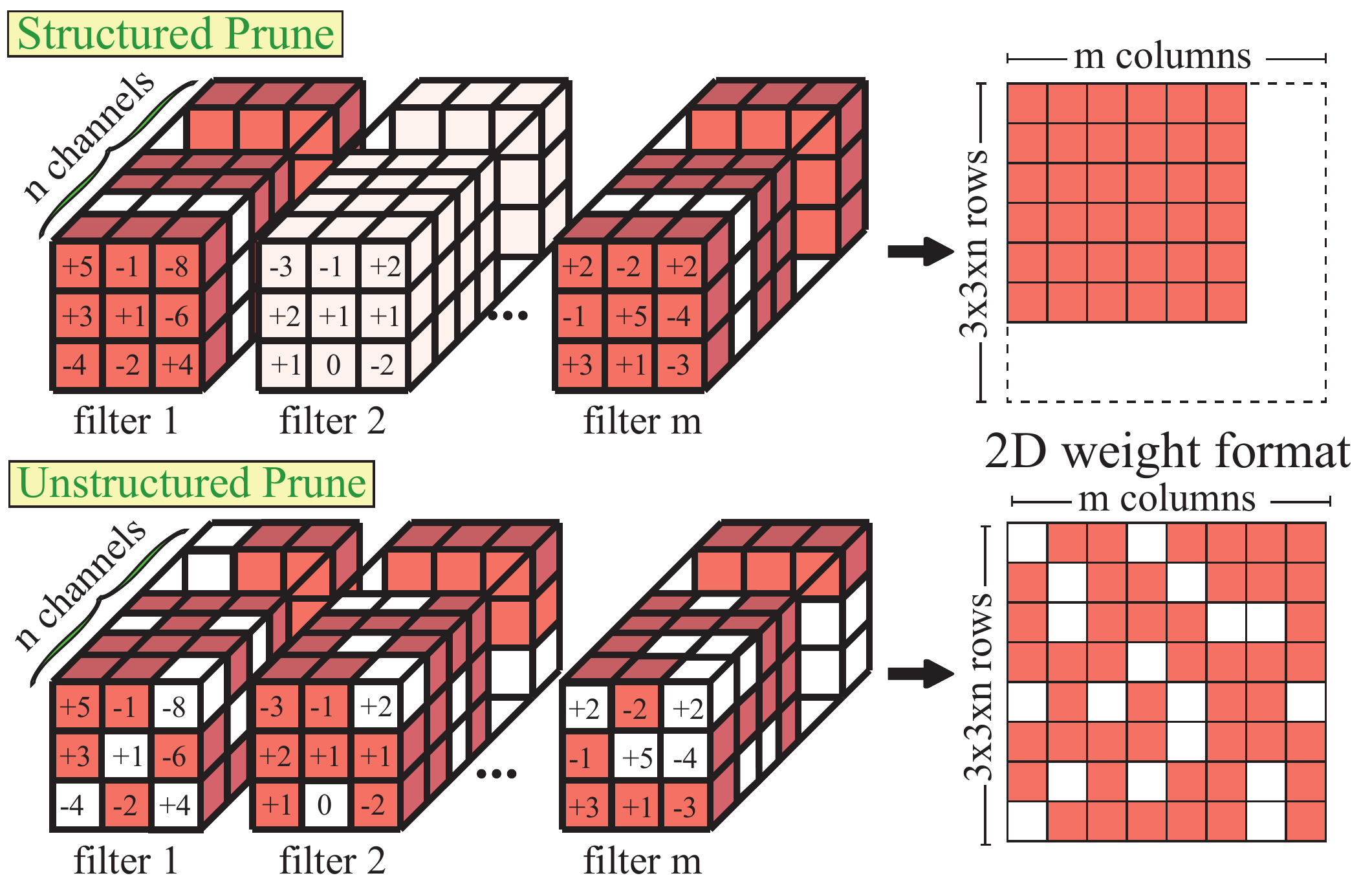}     
     \caption{Example of structured pruning and unstructured pruning.}
     \label{fig:structured}
\end{figure}

DNN Weight pruning~\cite{prune_survey,ren2018admmnn,9073635,li20206, ma2019nonstructured}, as one of the DNN model compression techniques, has also been investigated to reduce the weight storage and improve performance for ReRAM-based acceleration designs~\cite{yuan2019ultra, Ma_2020}.
The weight pruning consists of two main categories: 1) structured pruning and 2) unstructured pruning, as shown in Figure~\ref{fig:structured}.

\subsubsection{Structured pruning}
\textbf{Structured pruning} prunes the entire channel(s)/filter(s) of DNN weights~\cite{wen2016learning,he2017channel}. After removing the pruned weights, the size of weight matrix can be reduced in a structured manner and a regular shape of the weight matrix can be maintained.
This makes structured pruning hardware friendly.
By applying structured pruning to ReRAM-based DNN accelerator design, the number of ReRAM crossbars and the corresponding peripheral circuits can be reduced and hence an effective reduction of hardware cost in power and area.
However, due to the coarse pruning granularity of structured pruning, the pruned DNN models usually suffer from considerable accuracy loss.

\subsubsection{Unstructured pruning}

\textbf{Unstructured pruning} adopts a fine-grained pruning strategy, which prunes the weights at arbitrary locations~\cite{han2015deep,frankle2018lottery}.
This ensures higher flexibility to find the optimized pruned location, which is beneficial to preserve a higher model accuracy.
However, in order to accelerate the computation by leveraging unstructured sparsity (i.e., zeros in an irregular shape), it is inevitable to introduce additional indices to locate the non-zero weights during the computation.
It is generally considered non-hardware friendly. 
Thus, hardware designs, such as ReRAM-based accelerator, cannot take advantage of unstructured pruning to achieve hardware reduction and computation acceleration~\cite{ma2019nonstructured}.

\section{The Impact of Defect on DNN}
\label{sec:3}
Previous studies have shown the promising features and performances of ReRAM-based crossbars to be used in DNN accelerations. 
And compared to the conventional CMOS-based designs, ReRAM-based design achieves a great advantage in terms of power and area and is especially suitable for IoT devices.
However, the defect (e.g., stuck-at-fault defect) caused by the hardware imperfection during the manufacturing process or aging with time is still one of the key challenges that prevent the ReRAM to become a practical technology in computing and memory devices.

\begin{figure} [b]
     \centering
     \includegraphics[width=0.8\columnwidth]{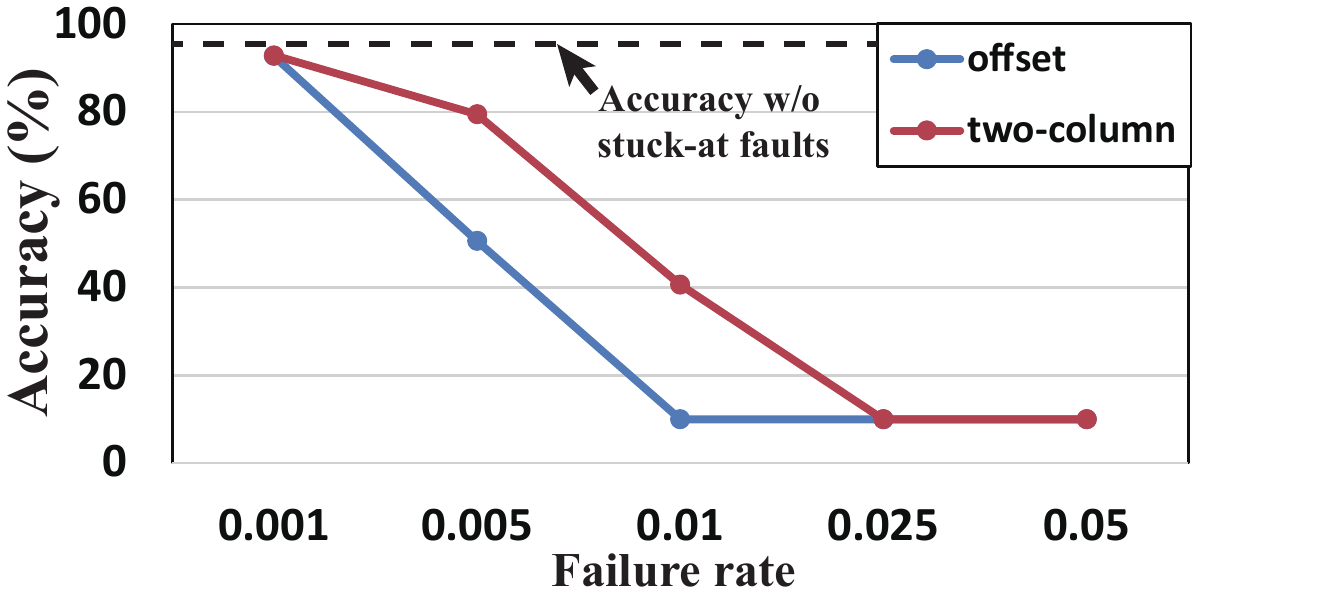}     
     \caption{Impact of stuck-at faults on accuracy using ResNet18 on CIFAR10.}
     \label{fig:offset_two_col}
\end{figure}

For example, Figure~\ref{fig:offset_two_col} shows the accuracy degradation caused by stuck-at-fault defects on ResNet18 using CIFAR10 dataset.
The defect model reported in~\cite{failure_model} is used. 
Note that the value of DNN weights can be either positive or negative, but the weights stored in ReRAM crossbars are represented by the conductance value of ReRAM cells, which can only be positive values.
Thus, there are two main mapping methods to make sure the results can be calculated correctly. 
The general way is to decompose each weight into a positive portion and a negative portion, then map them to two separate ReRAM crossbar columns.
Another way is to add an offset to the original weights when mapping them to the crossbars. This will shift all negative weights to the positive range. 
By doing so, with a certain cost of offset circuitry, half of the crossbars can be saved compared to the two-column mapping method. 
As the Figure~\ref{fig:offset_two_col} shows, both mapping methods suffer severe accuracy loss compared to the ideal case, where the offset method shows much worse tolerance to the defect.

The current solutions to handle the stuck-at-fault defects (e.g., defect-aware retraining, crossbar permutation, utilizing redundant crossbars) require an optimization process for each individual device (chip). 
Obviously, the cost of retraining and remapping for each device is too high, and it is unpractical for the mass-produced IoT products.
In addition, as IoT devices are used, new defects may occur, which may significantly mitigate the impact of the optimization methods.

\begin{quote}
\textbf{Question 1}: Is there a way that can improve the tolerance of stuck-at-fault defects uniformly instead of handling each IoT device individually?
\end{quote}

Weight pruning technique, which creates zeros weights in DNN models, might be a potential solution for this question.
This feature seems to be naturally beneficial to improve the stuck-off tolerance of DNN models, because when mapping the weights on to the ReRAM crossbars, some location of stuck-off faults may overlap with the weights that have been already pruned to zeros during the training (pruning) process on software.
The unstructured pruning is usually considered non-hardware friendly~\cite{ma2019nonstructured} and is not adopted in hardware designs.
However, for our case, the unstructured pruning would be a better choice than structured pruning, because unstructured pruning can create more zeros while maintaining a higher accuracy.

\begin{quote}
\textbf{Question 2}: Can the unstructured pruning be used in hardware design to improve tolerance of stuck-at-fault? If true, what is the relation between the pruning ratio and the model's fault tolerance?
\end{quote}

However, as reported in~\cite{failure_model}, the failure rate of stuck-on fault is as 5.2$\times$ higher than stuck-off fault. This will make the stuck-on fault become the most severe issue and hinder the overall tolerance improvement achieved by pruning (that only affects stuck-off fault).

\begin{quote}
\textbf{Question 3}: Is there an efficient way to improve the overall fault tolerance when stuck-on fault is the major issue?
\end{quote}

In the following paper, we will answer the above three questions.

\section{Improving DNN Fault Tolerance using Weight Pruning}

Stuck-at-faults is one of the main factors that causes the mismatch between software trained model and actual model mapped onto the hardware (e.g., ReRAM crossbar). 
This mismatch usually leads to undesirable performance degradation (i.e., accuracy drop) on hardware implementations.
To answer the Question 1 and Question 2, we investigate using the weight pruning technique to improve the DNN tolerance to the stuck-at fault without conducting optimizations for each individual device.

\subsection{Problem Formulation}

We define the occurrence of mismatch between
a trained weight $W$ and its actual value represented by the ReRAM crossbar $W_{act}$ as:
\begin{equation}
\label{equ1}
    {v(w, w_{act})}=
\begin{cases}
     0 & w_{act}=w, \\ 
 1 & w_{act}\neq w.
\end{cases}
\end{equation}

Given the failure rate of stuck-off fault and stuck-on fault of a ReRAM crossbar as $P_{sa0}$ and $P_{sa1}$, 
then the expectation of the occurrence of a DNN weight mapped on a ReRAM crossbar that is different to the trained value can be represented as:
\begin{equation}
\label{equ2}
E=0\cdot p(v=0)+1\cdot p(v=1)=P_{sa0}+P_{sa1}.
\end{equation}

However, when incorporating the weight pruning, the locations of the stuck-off fault may overlap with the pruned weights (0s), then the mismatch is avoided.
Considering the pruning ratio of a DNN $R_{p}=\frac{\#\_of\_zero\_weights}{total\_\#\_of\_weights}$, then the expectation of the occurrence of weight mismatch becomes:
\begin{equation}
\label{equ3}
E^{\prime}=P_{sa0}\cdot (1-R_p)+P_{sa1}.
\end{equation}
Obviously, the $E^\prime$ is smaller than $E$, which indicates that the weight pruning can statistically reduce the weight mismatch caused by stuck-off fault and can potentially mitigate the accuracy degradation.

\subsection{Exploring the Effects of Weight Pruning on DNN Stuck-at Fault Tolerance}
As we discussed above, the weight pruning seems an effective technique can be used to reduce the weight mismatch caused by stuck-off fault. Theoretically, a higher prune ratio leads to a higher probability that the stuck-off fault location overlaps with the pruned weight.
However, for real DNN models, the accuracy degradation caused by the stuck-off fault is not decreasing monotonically as the prune ratio increasing.

\begin{figure} [tbhp]
     \centering
     \includegraphics[width=1\columnwidth]{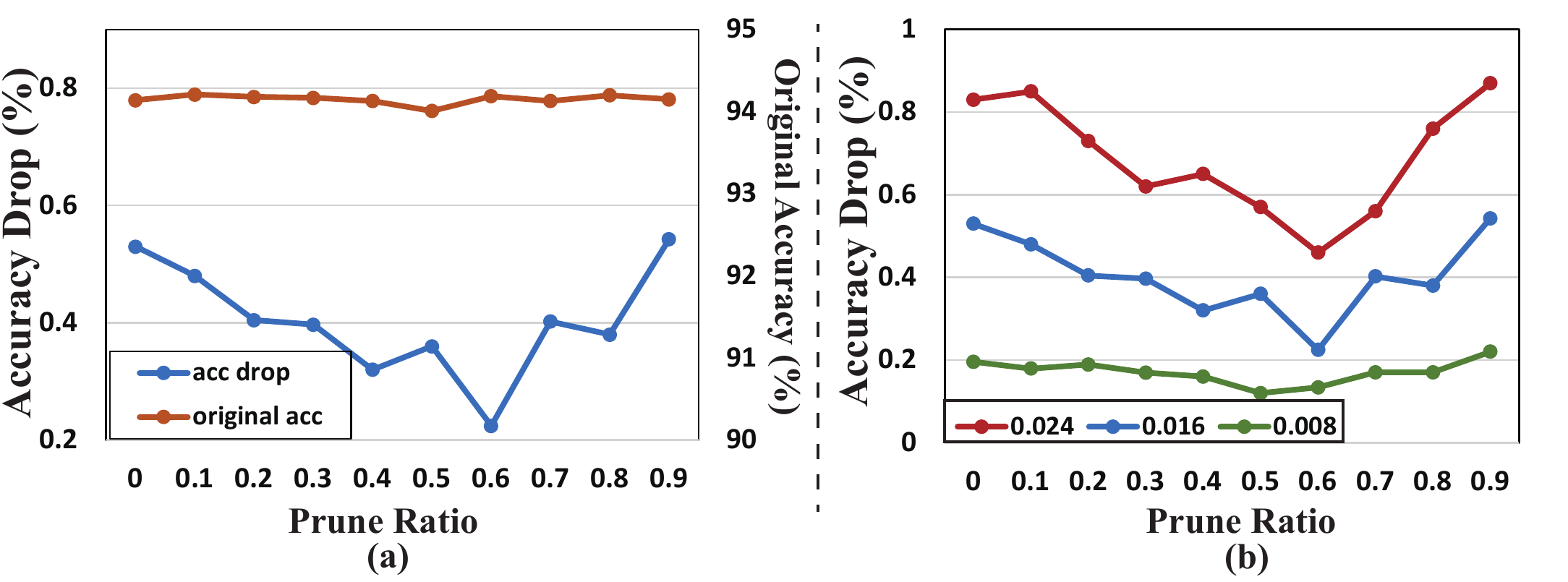}    
     \caption{Results on CIFAR10 dataset using ResNet18 under different pruning ratios: (a) Original model accuracy and accuracy drop caused by the stuck-off fault (0.016 fault rate), and (b) comparison of accuracy drop under different rate of stuck-off fault.}
     \label{fig:sa0_cifar}
\end{figure}

The Figure~\ref{fig:sa0_cifar} shows the results of accuracy drop caused by stuck-off fault under different unstructured pruning ratio on CIFAR10 dataset using ResNet18 models.
As shown in Figure~\ref{fig:sa0_cifar} (a), we evaluate the models with the prune ratio from 0 (i.e., not prune) to 0.9 (i.e., prune 90
\% weights). The original accuracy represents the model accuracy after pruning without incorporating stuck-off fault. 
We can observe that even 90\% of weights are pruned, the model can still retain a high accuracy (similar to the unpruned model) due to the high model redundancy on the CIFAR10 dataset and high flexibility of unstructured pruning.
We can also observe that as the pruning ratio increases, the accuracy drop caused by the stuck-off fault decreases and reaches its minimum value at the pruning ratio of 0.6, then the accuracy drop goes up again.

The reason is that, under a relative low pruning ratio, a higher weight pruning ratio can create more zero weights in the DNN model, which is possible to let more stuck-off fault locations overlap with those pruned weights (0s) and hence a less accuracy drop caused by the mismatch between trained weights and their actual mapped values.
However, when the pruning ratio is relatively high, few weights are left in the model, and each of them becomes very important.
Thus, the robustness of the model is reduced, and the faults that hit the non-zero weights will lead to more significant damage to the accuracy.
A similar trend can also be observed in Figure~\ref{fig:sa0_cifar} (b), which shows the comparison of accuracy drop under different stuck-off fault rates. 
We can find that weight pruning is more effective in improving fault tolerance under a higher fault rate.

\begin{figure} [tbhp]
     \centering
     \includegraphics[width=1\columnwidth]{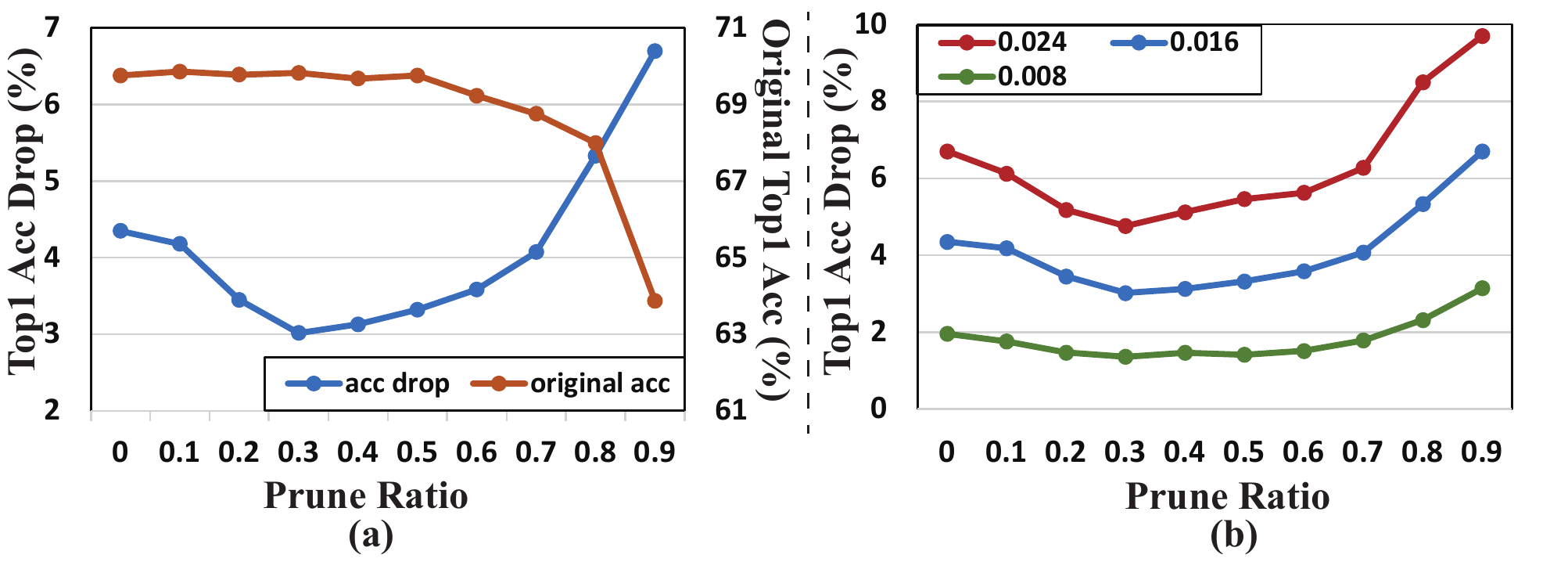}     
     \vspace{-1.5em}
     \caption{Results on ImageNet dataset using ResNet18 under different pruning ratios: (a) Original model accuracy and accuracy drop caused by the stuck-off fault (0.016 fault rate), and (b) comparison of accuracy drop under different rate of stuck-off fault.}
     \label{fig:sa0_imagenet}
\end{figure}

Figure~\ref{fig:sa0_imagenet} shows similar comparisons on ImageNet dataset using ResNet18.
Since the image classification task on ImageNet dataset is much complicated than on CIFAR10 dataset, the model is more sensitive to the stuck-at faults. 
We can observe that by using weight pruning, the accuracy drop caused by stuck-off fault can be mitigated by up to 2\% (Top-1 error) for the case under a 0.3 pruning ratio with a 0.024 fault rate compared with 0 pruning ratio case.

\subsection{A Hierarchical Progressive Pruning Method}
In order to mitigate the accuracy drop caused by stuck-at faults without losing network robustness, we propose a hierarchical progressive pruning method, which can find the best-suited pruning ratio for DNNs. Our method is derived by considering two key observations:
1) as the pruning ratio increases, the accuracy drop caused by stuck-at-fault defects will first decrease until reaching a minimum point, then increase again;
2) Larger DNN layers (i.e., usually the later layers in a neural network) have higher redundancy and can be pruned with higher pruning ratio without compromising accuracy.

The overall flow of our hierarchical progressive pruning method is shown in Algorithm~\ref{alg:hierachical}.
We first divide a given DNN model into several blocks according to the layer sizes.
Thus, each block contains several layers that have similar layer size (e.g., layers with 64 channels).
At the beginning, all the blocks are in the active model block list $B$.
We apply a small starting pruning ratio $p$ and get the pruned model $M$.
Note that, for this step, any pruning algorithm can be used to get the pruned model. 
And we use unstructured pruning in our method since, in this paper, we focus on creating more zeros rather than reducing hardware cost.
Then, we add stuck-at-fault defects on the model $M$ and evaluate the model accuracy with fault defect, and repeat many times to get the average accuracy.
If the accuracy performs better (or worse but within a threshold) then, we increase the pruning ratio and run test the accuracy again.
Otherwise, we remove the first block (which has the smallest block size) from $B$ and test the accuracy with the same pruning ratio again.
In this way, we can gradually increase the pruning ratio while narrow down the selected blocks.
As a result, we can obtain a pruned model with different pruning ratios in block-wise while delivering the best fault tolerance.

\begin{algorithm}[t]
    \textbf{Initialization:} \\ $M_{init}:$ well-trained model, \\ $th:$ threshold of acceptable accuracy drop, \\ Model blocks $B=[b_1, b_2, b_3, \dots, b_n]$, \\ $R_{range} = [r_1, r_2, r_3, \dots , r_m]$ candidate pruning ratios in ascending order, \\ Current model $M_{cur} = M_{init}$, \\ Best model $M_{best} = M_{init}$ \;
    \BlankLine
    \For{$p$ in $R_{range}$}{
    Prune blocks $B$ in $M_{cur}$ with prune ratio $p$\;
    Get pruned model $M$\;
        \For{num of test runs}{
              Add stuck-at-fault defects on model $M$\;
              Evaluate  model accuracy\;
        }
    Get average model accuracy $Acc(M)$\;
    \uIf{$Acc(M_{cur}) - Acc(M) < th$}{
        \If{$Acc(M) > Acc(M_{cur}$)}{
            $M_{best} = M$\;
        }
        $M_{cur} = M$\;
    }
    \Else{$B.pop\_front()$ \quad // remove first block in $B$\;}
    \If{$B$ is empty}{break\;}
    }
    \Return $M_{best}$  \quad  // model with highest accuracy
    \caption{Hierarchical Progressive Pruning.}
    \label{alg:hierachical}
\end{algorithm}

\subsection{Differential Mapping Scheme}
By incorporating the weight pruning, the accuracy drop caused by stuck-off fault can be mitigated.
However, the performance of hardware implementation needs to consider both the stuck-off fault and stuck-on fault.
Moreover, according to~\cite{failure_model}, 
the failure rate of stuck-on fault can be 5.2$\times$ higher than the failure rate of stuck-off fault. 
In this case, the stuck-on fault becomes the major issue leading to the accuracy drop rather than the stuck-off fault.
To solve this problem, we propose a differential ReRAM crossbar mapping scheme to improve the network tolerance to stuck-on fault.

The traditional mapping scheme decomposes the weight into a positive magnitude portion $w^{+}$ and a negative magnitude portion $w^{-}$,
and uses two ReRAM cells to represent positive and negative portions separately.

Different from the traditional mapping scheme that decomposes the weight into a positive magnitude portion $w^{+}$ and a negative magnitude portion $w^{-}$
and sums them to reconstruct the original weight value during the computation, 
our proposed differential mapping scheme represents the weight value by using the difference between two ReRAM cells.

Given a weight value $w$ scaled to the range [-1,1], then it will be mapped as:

\begin{minipage}{0.47\linewidth}
\small
\begin{eqnarray*}w_a=
\begin{cases}
 1 & w\geq 0, \\
 1-|w| &  w<0,
\end{cases}
\end{eqnarray*}
\end{minipage}
\begin{minipage}{0.47\linewidth}
\small
\begin{eqnarray*}w_b=
\begin{cases}
 1-w & w>0, \\
 1 &  w\leq 0,
\end{cases}
\end{eqnarray*}
\end{minipage}

where the $w_a$ and $w_b$ are the two ReRAM cells (with normalized range to [0,1]) used to represent the weight $w$. During the computations, the original weight value is obtained by using $w_a - w_b$.

Our differential mapping scheme ensures a more number of 1s to be mapped on the ReRAM crossbars.
There is always at least one 1 to be mapped on one of the two ReRAM cells. 
And both of the ReRAM cell will be 1 when weight value is zero.
In this way, the weight pruning will create more 1s, which will increase the opportunity that the stuck-on faults overlap with those 1s and mitigate the accuracy drop caused by the stuck-on fault.
It can improve the performance significantly, since the failure rate of stuck-on fault is higher than stuck-off fault and will lead to more significant accuracy drop, although the differential mapping scheme only improve the tolerance to the stuck-on fault.

\begin{figure} [!tb]
     \centering
     \includegraphics[width=1\columnwidth]{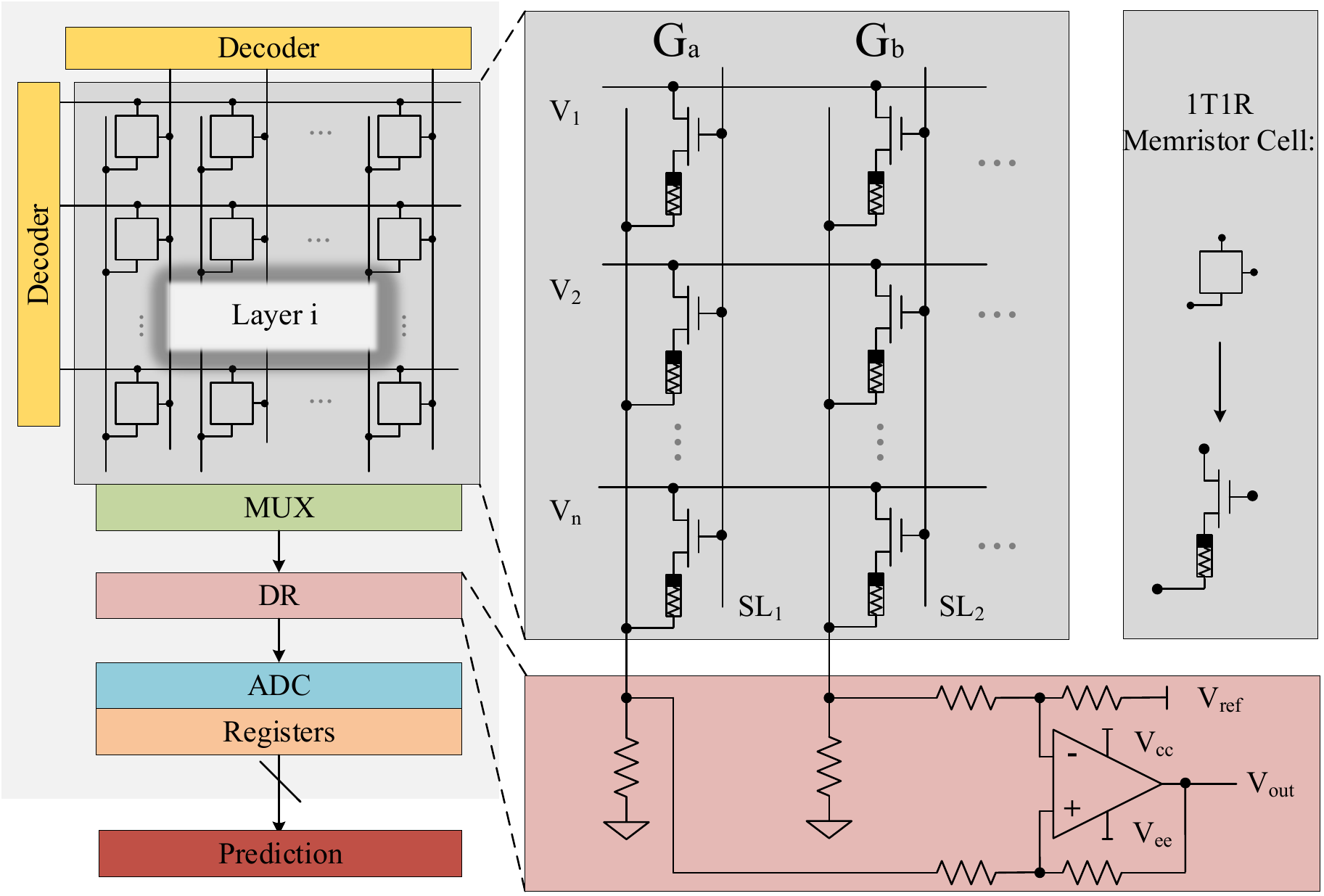}    
     \caption{Hardware implementation of proposed differential mapping scheme.}
     \vspace{-1.5em}
     \label{fig:dm}
\end{figure}

The Figure~\ref{fig:dm} shows the hardware implementation of our proposed differential mapping scheme. 
Compared with traditional hardware implementation~\cite{double_mapping} that weight values are calculated by two ReRAM columns and an OpAmp-based subtractor, the differential mapping method still utilizes similar arithmetic circuit without increasing hardware cost~\cite{double_mapping2}.
When performing differential mapping, the $w_a$ and $w_b$ values are mapped to the ReRAM cells in the $G_a$ and $G_b$ columns, respectively. 
When reading the weight value, the current sum of the $G_a$ and $G_b$ columns is subtracted by the arithmetic circuit to obtain an effective activation value.

\section{Evaluation}
\subsection{Experimental Setup}
We evaluate our methods on CIFAR10 dataset and ImageNet dataset using ResNet18 for image classification task.
We also validate our methods on YOLO-v4 for object detection task using MS COCO dataset.
The ReRAM failure model is adopted from~\cite{failure_model} with the ratio of stuck-off and stuck-on fault is 1:5.2.
The ADMM-based pruning algorithm~\cite{ zhang2018systematic} is used for weight pruning during the hierarchical progressive pruning process.
All model training, pruning, and accuracy evaluations are conducted on a GPU server using PyTorch API. Each accuracy result is obtained by averaging the results of 100 runs.

\subsection{Comparison Results}
\begin{figure} [!tb]
     \centering
     \vspace{-1.2em}
     \includegraphics[width=1\columnwidth]{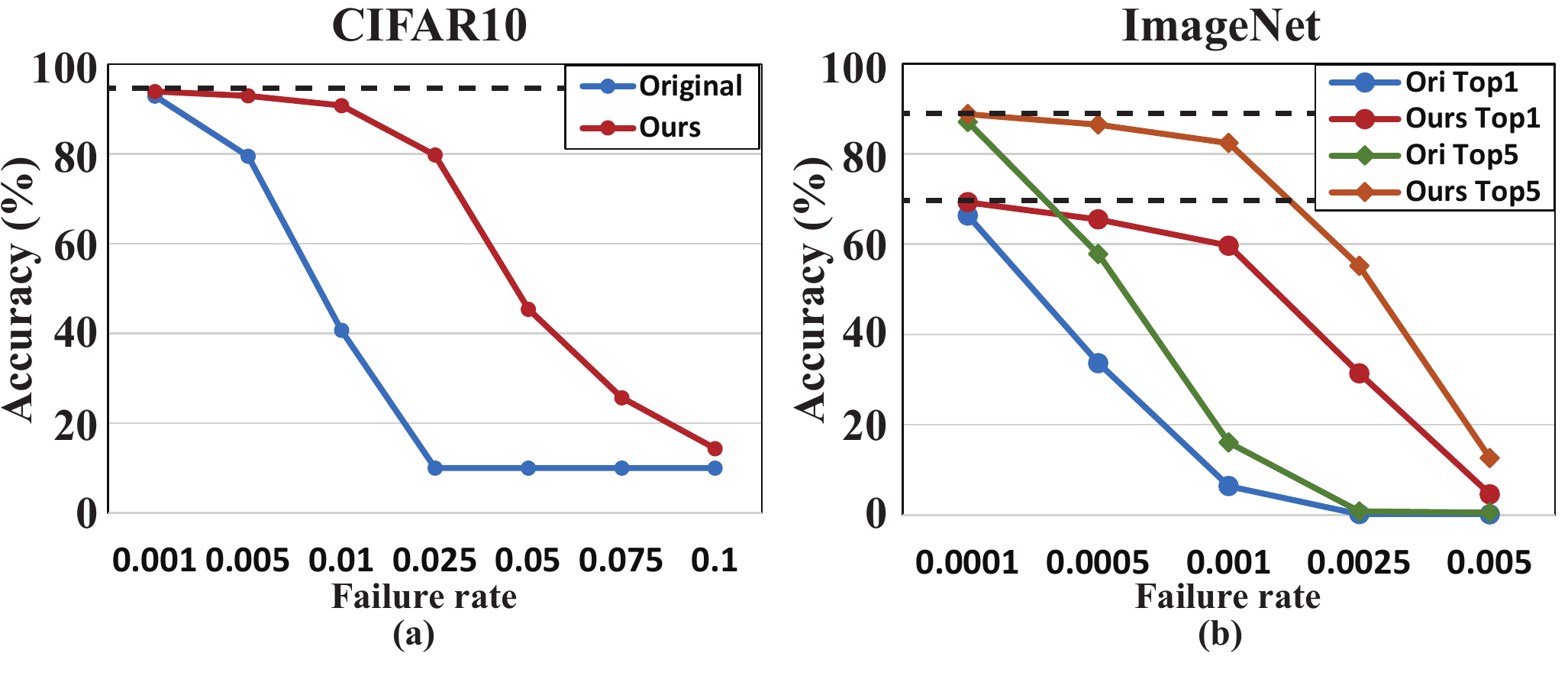}  
     \caption{Comparison results on CIFAR10 dataset and ImageNet using ResNet18 under different failure rate.}
     \vspace{-1em}
     \label{fig:cifar_imagenet}
\end{figure}

We first evaluate the model fault tolerance via model accuracy on image classification task.
As shown in Figure~\ref{fig:cifar_imagenet}, we compare the model fault tolerance results optimized by our methods (including unstructured pruning and differential crossbar mapping) to the original unpruned model with traditional two-column mapping (as discussed in Section~\ref{sec:3}).
We do not compare to the offset mapping scheme since the fault tolerance of the offset scheme is worse than the traditional two-column scheme.
On CIFAR10 dataset, the original model accuracy is 94.1\% without introduce the stuck-at faults. Under failure rate of 0.001, our method only has 0.2\% accuracy drop on average, where the accuracy drop of the traditional mapping is 1.2\%.
It can be observed that a severe accuracy drop occurs to the traditional mapping under failure rate of 0.005, where our optimizations can preserve a high accuracy under a failure rate of 0.01.
For the ImageNet, since the classification task on ImageNet is harder than CIFAR10, the network is more sensitive to the stuck-at faults. As we can see, our optimizations clearly provides a better fault tolerance than the traditional two-column mapping.

\begin{figure} [!tb]
     \centering
     \includegraphics[width=0.9\columnwidth]{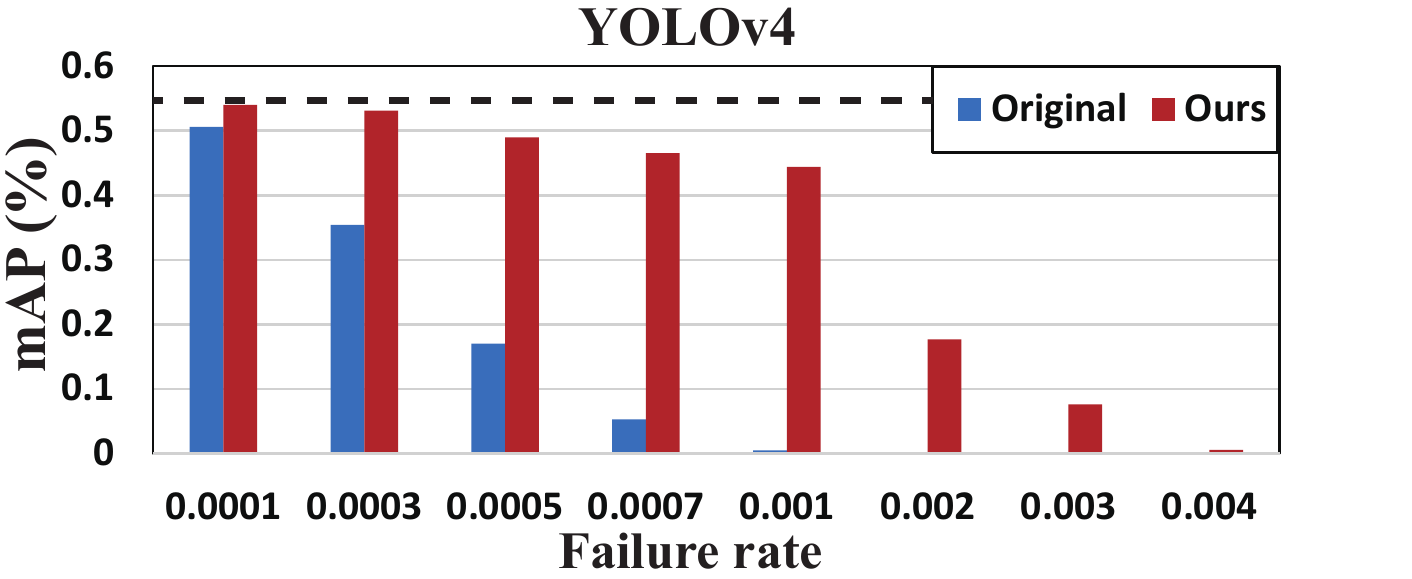}  
     \vspace{-0.8em}
     \caption{Comparison results of YOLOv4 on MS COCO dataset under different failure rate.}
     \vspace{-1.5em}
     \label{fig:yolo}
\end{figure}

We also validate our methods on object detection task.
Figure~\ref{fig:yolo} shows the comparison results on MS COCO dataset using YOLOv4 for object detection task.

All the results shows that our method can tolerate almost an order of magnitude higher failure rate than the traditional two-column method, which demonstrates the effectiveness of our optimizations.

\section{Conclusion}
In this paper, we rethink the value of weight pruning in ReRAM-based DNN design from the perspective of model fault tolerance.
We 
propose a hierarchical progressive pruning method, which can find the best-suited pruning ratio for DNNs to improve the model fault tolerance.
And a differential mapping scheme is proposed to improve the fault tolerance under a high stuck-on fault rate.

\section*{Acknowledgement}
This work is funded by the National Science Foundation Awards CCF-1937500 and CNS-1909172.


\end{document}